\documentclass{article}

\usepackage{PRIMEarxiv}

\usepackage[utf8]{inputenc} 
\usepackage[T1]{fontenc}    
\usepackage[numbers,sort&compress]{natbib} 
\usepackage{xcolor}
\usepackage{hyperref}       
\hypersetup{colorlinks=true, linkcolor=blue!60!black,
  citecolor=blue!60!black, urlcolor=blue!60!black}
\usepackage{url}            
\usepackage{booktabs}       
\usepackage{amsfonts}       
\usepackage{amsmath}        
\usepackage{amssymb}        
\usepackage{amsthm}         
\newtheorem{definition}{Definition}
\usepackage{algorithm}       
\usepackage{algpseudocode}   
\usepackage{nicefrac}       
\usepackage{microtype}      
\usepackage{fancyhdr}       
\usepackage{graphicx}       
\graphicspath{{media/}}     

\title{Weight-Space Programming of Policies and Their Verification via Model Checking
}

\author{
  Dennis Gross \\
  \texttt{dennis@artigo.ai} \\
}

\begin{document}
\maketitle

\begin{abstract}
Large language models (LLMs) can write the weights of small neural networks
directly, without gradient descent.
This weight-space programming creates policies for sequential decision-making from a task description alone, but
nothing guarantees that a written policy behaves as intended. We combine
weight-space programming with probabilistic model checking. Given an
environment modeled as a Markov decision process and a requirement in
probabilistic computation tree logic (PCTL), an LLM writes all weights of a
policy network, optionally repairs them on a dataset of optimal actions, and
receives model-checking feedback until the requirement holds or a round limit
is reached. The model checker
Storm computes, rather than estimates from simulations, the probability that
the policy satisfies the requirement, so every returned policy is verified. With two LLMs on three
environments, the created policy of each of the 18 runs attains the optimal
probability, before any dataset or model-checking feedback. Four of six policies
programmed for two transporter jobs also complete three jobs with probability 1,
so they generalize to a task that the LLM never saw. The pipeline also programs
a policy that satisfies a complex safety requirement expressed with the PCTL until
operator in all six runs. In one of them, dataset repair raised the accuracy to 98.4\% but
broke the requirement, and one round of model-checking feedback restored it.
\end{abstract}


\section{Introduction}
In \emph{reinforcement learning (RL)}, an agent learns a policy by interacting with an
environment and collecting rewards that steer it toward a fixed
objective~\cite{mnih2015human}.
Such policies are
typically neural networks trained through \emph{gradient descent}, which repeatedly
nudges each weight in the direction that increases the expected reward.
\emph{Weight-space programming} offers a different path: a \emph{large language model
(LLM)} reads and writes a network's weights directly, guided only by a
natural-language specification~\cite{takehi2026sparks}.
Although the idea is still new, early results show that LLMs can write small policies from scratch
and repair failing ones by editing a small fraction of their weights~\cite{takehi2026sparks}.
Two questions remain open. 
First, LLMs make mistakes, so nothing guarantees that
a written policy behaves as intended~\cite{huang2024survey}.
Second, it is unclear whether
weight-space programming can encode complex safety requirements at all.

These complex requirements go beyond the scalar rewards that RL maximizes~\cite{DBLP:journals/aamas/VamplewSKRRRHHM22}. Consider a
robot that must reach its goal without ever colliding along the way. In
\emph{probabilistic computation tree logic (PCTL)}~\cite{hansson1994logic}, the
\emph{until} operator states this precisely as
$P_{\geq 0.99}(\neg \mathit{collision}\ U\ \mathit{goal})$: with
probability at least $0.99$, the robot avoids collisions \emph{until} it reaches
the goal.
A reward function can penalize collisions, but it only trades them off
against progress and cannot enforce such a probability bound. In general, not
every such requirement can be expressed as a scalar
reward~\cite{DBLP:journals/aamas/VamplewSKRRRHHM22,gross2022coolmc}. \emph{Probabilistic model checking}~\cite{baier2008principles}
via the tool COOL-MC~\cite{gross2022coolmc} fills this gap.
Given an environment modeled in the PRISM language~\cite{kwiatkowska2011prism} and a policy, it
constructs the Markov chain the policy induces and uses the model checker
Storm~\cite{hensel2022probabilistic} to compute, rather than estimate, the probability that a PCTL property holds.
This makes it a natural judge for LLM-written policies.

In this paper, we combine weight-space programming with probabilistic model
checking. Our pipeline has three steps, none of which uses gradient descent.
First, an LLM receives the PRISM model, the PCTL property, a short
description of the environment, and the shape of the policy network, and writes
all of its weights. With the weights, it lists sample states and the action it
intends for each, and we run the network on them to catch faulty weights early.
Second, if a dataset of optimal actions is available, the LLM sees the states
where its network disagrees with the dataset and edits the weights. Third,
COOL-MC model checks the policy. If the property is violated, the LLM
receives the computed probability, sample states with the network's actions,
and, if an earlier network was better, that network, and edits the weights again. The second step
repeats until the accuracy on the dataset reaches a target, the third until the
property is satisfied, each up to a round limit.
Because Storm checks every policy that the pipeline returns, a returned policy
that violates the property never goes unnoticed.

Across two LLMs (DeepSeek-V4-Pro~\cite{xu2026deepseek} and Kimi-K3.1-Pro~\cite{team2026kimi}) and three PRISM environments, the created policy of each of the
18 runs reaches the optimal probability. The networks of Kimi-K3.1-Pro select an
optimal action in 98.9\% of the dataset states on average after creation and in
all of them after dataset repair. Policies programmed for two transporter jobs
also complete three jobs with probability 1 in 4 of 6 runs, so they generalize
to a task that the LLM never saw.
For a complex fuel-reserve safety
requirement, the pipeline finds a satisfying policy in all 6 runs; in one of
them, model-checking repair restores the requirement after dataset repair had
broken it.
The most common generation error is a weight matrix with a wrong
number of rows, which the validation catches.

Our \textbf{main contributions} are:
\begin{itemize}
  \item a pipeline in which an LLM writes and repairs the weights of a neural
        network policy, guided by a self-check and by probabilistic model
        checking;
  \item an evaluation of two LLMs on three PRISM environments that compares the
        written policies after each step with the optimal actions and
        probabilities computed by Storm and counts how often the LLMs produce
        faulty weights; and
  \item an analysis of whether the written policies generalize from two to
        three transporter jobs and whether the pipeline can program a complex
        safety requirement expressed with the until operator.
\end{itemize}
The rest of this paper is structured as follows.
Section~\ref{sec:related} reviews related work, and
Section~\ref{sec:background} introduces probabilistic systems, probabilistic
model checking, LLMs, and weight-space programming.
Section~\ref{sec:method} presents our pipeline and its limitations, and
Section~\ref{sec:experiments} evaluates it on four research questions.
Section~\ref{sec:discussion} discusses its benefits and future applications,
and Section~\ref{sec:conclusion} concludes the paper and outlines future work.

\section{Related Work}\label{sec:related}

\paragraph{Weight-space learning.}
A growing body of work treats neural network weights as
data~\cite{han2026survey}.
Hypernetworks generate the weights of another
network~\cite{DBLP:journals/corr/HaDL16},
equivariant architectures process weights while respecting their permutation
symmetries~\cite{navon2023equivariant,zhou2023permutation},
and generative models synthesize new weights from collections of trained
networks~\cite{peebles2022learning} or LoRA adapters from natural-language task
descriptions~\cite{charakorn2025text}.
Many of these generators, however, largely reproduce their training checkpoints~\cite{zeng2026generative}.
All of these methods must be trained for their purpose, several of them on
collections of trained networks, whereas we use a general-purpose LLM without
such training.

\paragraph{Editing and programming weights.}
Weights can also be changed directly.
Task arithmetic adds and subtracts task vectors, the differences between the
weights of fine-tuned and pre-trained models~\cite{ilharco2022editing},
and ROME edits factual associations in language models through rank-one
updates~\cite{meng2022locating}.
Tracr compiles programs written in the RASP
language~\cite{weiss2021thinking} into transformer
weights~\cite{NEURIPS2023_771155ab}, but requires such a formal program as
input.
Weight-space programming removes this need: an LLM writes weights from a
natural-language specification~\cite{takehi2026sparks}.
Follow-up work conjectures that LLMs can also improve image-restoration
diffusion models by editing their weights~\cite{zheng2026weight}.
Orbit audits test whether weight readers, including LLMs, recognize the
function that weights encode, by applying neuron permutations and rescalings
that leave this function unchanged~\cite{cha2026consistency}.
These works evaluate weights empirically, for instance, by game score or
recognition accuracy, or conjecture improvements without
experiments~\cite{zheng2026weight}. We instead check written policies against PCTL
properties. Because model checking judges the function a network computes,
its verdict is unaffected by such reparameterizations.

\paragraph{LLMs for policy and reward synthesis.}
LLMs have been used to write policies as code~\cite{liang2023code}
and to design reward functions for RL~\cite{ma2024eureka}.
Programmatic policies can also be easier to interpret and verify than neural
ones~\cite{verma2018programmatically}.
In contrast, we keep the policy a neural network and let the LLM write its
weights directly, so that existing verification tools for neural policies apply.

\paragraph{Safe RL and verification.}
Safe RL constrains the behavior of agents, for example, by bounding the
worst-case cost during training~\cite{Yang2022safety-constrained},
by shielding unsafe actions~\cite{brorholt2026adaptive,Konighofer2025Shields},
or by learning from temporal-logic objectives instead of hand-designed
rewards~\cite{DBLP:conf/tacas/HahnPSSTW19}.
Probabilistic model checking verifies trained policies against PCTL
properties~\cite{gross2022coolmc},
and a policy that is known to be safe can be imitated by a neural network
through behavioral cloning~\cite{gross2026formally} or represented and
generalized as a decision tree~\cite{ashok2021dtcontrol,azeem20251}.
These approaches obtain the policy through training or decision-tree learning,
or use LLMs as policies that are verified directly~\cite{gross2025verifying}.
We use model checking both to verify LLM-written policies and to give the LLM feedback for
repairing them.

\section{Background}\label{sec:background}
We introduce probabilistic systems, probabilistic model checking, large language
models (LLMs), and weight-space programming.

\subsection{Probabilistic Systems}\label{sec:bg-systems}
A \emph{probability distribution} over a set $X$ is a function
$\mu\colon X \rightarrow [0,1]$ with $\sum_{x \in X} \mu(x) = 1$. The set of all
distributions on $X$ is $\mathit{Distr}(X)$. We model the environment of a
sequential decision-making task as a Markov decision process
(Definition~\ref{def:mdp}).

\begin{definition}[MDP]\label{def:mdp}
A \emph{Markov decision process (MDP)} is a tuple
$M = (S, s_0, \mathit{Act}, \mathit{Tr}, \mathit{rew}, \mathit{AP}, L)$ where
$S$ is a finite, nonempty set of states; $s_0 \in S$ is an initial state;
$\mathit{Act}$ is a finite set of actions;
$\mathit{Tr}\colon S \times \mathit{Act} \rightarrow \mathit{Distr}(S)$ is a
partial probability transition function;
$\mathit{rew}\colon S \times \mathit{Act} \rightarrow \mathbb{R}$ is a reward
function; $\mathit{AP}$ is a set of atomic propositions;
$L\colon S \rightarrow 2^{\mathit{AP}}$ is a labeling function.
\end{definition}

Taking action $a$ in state $s$ leads to the next state $s'$ with probability
$\mathit{Tr}(s,a)(s')$. Because $\mathit{Tr}$ is partial, $\mathit{Tr}(s,a)$ may
be undefined, written $\mathit{Tr}(s,a) = \bot$, in which case $a$ cannot be
taken in $s$. The \emph{available actions} in a state $s \in S$ are thus
$\mathit{Act}(s) = \{a \in \mathit{Act} \mid \mathit{Tr}(s,a) \neq \bot\}$. The
labeling function assigns to each state the atomic propositions that hold in it,
such as $\mathit{goal}$ or $\mathit{collision}$. The reward function is used to
train policies with reinforcement learning (RL); verification does not need it.

We employ a \emph{factored state representation}, in which each state $s$ is a vector
of \emph{features} $(f_1, f_2, \ldots, f_d)$ with $f_i \in \mathbb{Z}$ for
$1 \leq i \leq d$, where $d$ is the state dimension. We specify MDPs in the
PRISM language~\cite{kwiatkowska2011prism}. The features are then the variables
of the PRISM model, with Boolean variables encoded as $0$ and $1$, and each
action has a \emph{guard}, a condition on the features that defines the states
in which the action is available.

\begin{definition}[Policy]
A \emph{memoryless deterministic policy} $\pi$ for an MDP $M$ is a function
$\pi\colon S \rightarrow \mathit{Act}$ that maps a state $s \in S$ to an
available action $\pi(s) \in \mathit{Act}(s)$.
\end{definition}

The policy is \emph{deterministic} because it selects exactly one action, and
\emph{memoryless} because this action depends only on the current state, not on
the states visited before. Throughout this paper, all policies are memoryless
and deterministic.

A \emph{discrete-time Markov chain (DTMC)} is a probabilistic system without
action choices: each state has exactly one distribution over next states.
Applying a policy $\pi$ to an MDP $M$ yields the \emph{induced DTMC} $D$, in
which all non-determinism, that is, the choice of actions, is resolved: in each
state $s$, the next state is drawn from $\mathit{Tr}(s, \pi(s))$. A \emph{path}
of $D$ is an infinite sequence of states $s_0 s_1 s_2 \ldots$ that starts in the
initial state and in which each state $s_{i+1}$ follows $s_i$ with positive
probability. A state is \emph{reachable} if a path visits it.

\subsection{Probabilistic Model Checking}\label{sec:bg-pmc}
Probabilistic model checking verifies whether a probabilistic system satisfies
a formal property~\cite{baier2008principles}. The general workflow with the
probabilistic model checker Storm~\cite{hensel2022probabilistic} is as follows.
First, the environment is modeled as an MDP, for instance, in PRISM. Next, the
informal \emph{requirements} on its behavior are formalized as
\emph{properties}. Given the MDP, a policy, and a property, Storm then checks
whether the induced DTMC satisfies or violates the property and computes the
probability that the property holds. Unlike simulation, which estimates this
probability from sampled paths, Storm computes it numerically on the whole
DTMC, up to a small numerical error. Among the most fundamental
properties are \emph{reachability properties}, which assess the probability of
reaching a particular set of states. For example, one might ask, ``Is the
probability of reaching an unsafe state less than $0.1$?''

Model checkers can proceed ``on the fly'', exploring only the parts of the DTMC
that are relevant to the verification. The tool COOL-MC~\cite{gross2022coolmc},
which connects RL policies with Storm, builds the induced DTMC this way:
starting in $s_0$, it queries the policy for an action in every reachable state
and adds only the next states of that action.

\paragraph{PCTL properties.}
We specify properties in probabilistic computation tree logic
(PCTL)~\cite{hansson1994logic}. A PCTL property has the form $P_{\sim p}(\rho)$,
where $\rho$ is a \emph{path property}, ${\sim} \in \{<, \leq, \geq, >\}$ is a
\emph{relational operator}, and $p \in [0,1]$ is a \emph{probability threshold}. The induced
DTMC $D$ \emph{satisfies} $P_{\sim p}(\rho)$ if the probability of the paths of
$D$ that fulfill $\rho$ meets the bound ${\sim}\,p$; otherwise, $D$
\emph{violates} it. Path properties are built from \emph{state formulas}, which
either hold or do not hold in a single state. In this paper, state formulas are
atomic propositions and conditions on the features, such as $f_1 = 2$, combined
with negation ($\neg$) and conjunction ($\wedge$); the state formula
$\mathit{true}$ holds in every state. Our experiments use two path properties,
which we describe on a high level below; for a detailed treatment, we refer
to~\cite{baier2008principles,hansson1994logic}.

\paragraph{Eventually path property.}
The \emph{eventually} operator $\Diamond$ expresses that a state formula $\varphi$ will
hold at some point in the future. A path fulfills $\Diamond \varphi$ if
$\varphi$ holds in some state of the path. Thus, $P_{\sim p}(\Diamond \varphi)$
states that the probability of eventually reaching a state where $\varphi$
holds meets the bound ${\sim}\,p$.

\paragraph{Until path property.}
The \emph{until} operator $U$ specifies that a state formula $\varphi$ holds
continuously until another state formula $\psi$ becomes true. A path fulfills
$\varphi\ U\ \psi$ if $\psi$ holds in some state of the path and $\varphi$ holds
in all states before it. Thus, $P_{\sim p}(\varphi\ U\ \psi)$ states that the
probability of reaching a state where $\psi$ holds, while $\varphi$ holds on the
way there, meets the bound ${\sim}\,p$. Eventually is the special case
$\Diamond \varphi \equiv \mathit{true}\ U\ \varphi$.

For example, consider a robot that must reach a goal without colliding. The
property $P_{\leq 0.01}(\Diamond\, \mathit{collision})$ requires that it
collides with probability at most $0.01$, and
$P_{\geq 0.99}(\neg \mathit{collision}\ U\ \mathit{goal})$ requires that it
reaches the goal without colliding on the way with probability at least $0.99$.

\paragraph{PCTL queries.}
Besides whether a property is satisfied, we are interested in the actual
probability. Although it is not formally permitted in PCTL, one can still use
Storm to compute the probability that a path property is fulfilled; we call this
a \emph{PCTL query}. In a slight abuse of notation, we write the query as
$P(\rho)$, omitting the bound, for the probability of the paths of $D$ that
fulfill $\rho$, and report its result as, for instance,
$P(\Diamond\, \mathit{goal}) = 0.82$: the policy eventually reaches the goal
with probability $0.82$. The query result also decides every property with the
same path property; for instance, the policy above satisfies
$P_{\geq 0.8}(\Diamond\, \mathit{goal})$ but violates
$P_{\geq 0.9}(\Diamond\, \mathit{goal})$. On the MDP $M$ itself,
$P_{\max}(\rho)$ and $P_{\min}(\rho)$ denote the maximal and minimal probability
over all policies. For the path properties we consider, memoryless deterministic
policies suffice to attain these values~\cite{baier2008principles}. They are
thus the best probabilities any policy can achieve, and we use them as the
reference for the policies that LLMs write.

\subsection{Large Language Models}\label{sec:bg-llm}
A \emph{token} $\tau$ is an atomic text element, such as a word, punctuation
mark, or symbol. A \emph{bag of tokens} $\Sigma$ is a finite, nonempty set of
tokens.

\begin{definition}[String]
A \emph{string} $\omega$ is a finite sequence of tokens chosen from some bag
$\Sigma$. The set of all strings over $\Sigma$ is $\Sigma^*$.
\end{definition}

\begin{definition}[Large-Language-Model Function]
An \emph{LLM function} is a total mapping
$f\colon \Sigma^* \rightarrow \mathit{Distr}(\Sigma^*)$ that assigns to every
input string $\omega \in \Sigma^*$, called the \emph{prompt}, a probability
distribution $f(\omega)$ over finite strings. A call of $f$ with $\omega$
returns one \emph{output string} $\omega'$ sampled from $f(\omega)$, written
$\omega' \sim f(\omega)$.
\end{definition}

An LLM generates its output string in an \emph{autoregressive} manner. Its neural
network, a \emph{transformer}~\cite{vaswani2017attention}, processes the text so far
and produces a probability distribution over the next token. A token is
stochastically selected from this distribution and appended to the text, which
is then re-fed into the transformer, until a designated stop token is selected.
This stochastic selection induces the distribution $f(\omega)$, so two calls
with the same prompt can return different output strings. LLMs can solve new tasks that are only described in the prompt,
including writing code~\cite{huynh2025large}, but their output strings are
not guaranteed to be correct.

\subsection{Weight-Space Programming}\label{sec:bg-wsp}
A \emph{policy network} is a feed-forward neural network
$N\colon \mathbb{Z}^d \rightarrow \mathbb{R}^{|\mathit{Act}|}$ that maps a state
$s = (f_1, \ldots, f_d)$ to one output $N(s)_a$ per action $a \in \mathit{Act}$.
With $k$ layers, it computes
\[
  N(s) = W_k h_{k-1} + b_k, \qquad h_0 = s, \qquad
  h_i = \mathrm{relu}(W_i h_{i-1} + b_i) \text{ for } 0 < i < k,
\]
where $\mathrm{relu}(z) = \max(0, z)$ is the \emph{rectified linear unit (ReLU)},
applied element-wise. The entries of $h_i$ are the \emph{neurons} of the $i$-th
hidden layer. We call the matrices $W_i$ and vectors $b_i$ together the
\emph{weights} of $N$. The policy network induces a policy $\pi$ that selects
the action with the largest output,
$\operatorname*{arg\,max}_{a \in \mathit{Act}} N(s)_a$. If this action is not
available in $s$, COOL-MC selects the first action of $\mathit{Act}(s)$ in
alphabetical order of the action names instead, so that
$\pi(s) \in \mathit{Act}(s)$ always holds.

Usually, the weights are found by gradient descent, which repeatedly adjusts
them to reduce a training loss or to increase the expected reward. In
weight-space programming~\cite{takehi2026sparks}, an LLM instead writes them
directly as numbers in its output string. This works because ReLU networks can
express rules exactly. Since features are integers, two neurons compute a \emph{step
function} for a
feature $f_j$ and a constant $c \in \mathbb{Z}$,
\[
  \mathrm{relu}(f_j - c + 1) - \mathrm{relu}(f_j - c) =
  \begin{cases} 1 & \text{if } f_j \geq c,\\ 0 & \text{otherwise,} \end{cases}
\]
and for $m$ values $g_1, \ldots, g_m \in \{0,1\}$, such as step functions, a
single neuron in the next layer computes their \emph{conjunction}
$\mathrm{relu}(g_1 + \cdots + g_m - (m-1))$, which is $1$ exactly if all of them
are $1$. A rule such as ``if $f_1 \geq 3$ and $f_2 \geq 2$, select action $a$''
thus needs four neurons in the first hidden layer (two step functions), one
neuron in the second hidden layer (their conjunction), and a large weight from
that neuron to the output $N(s)_a$. The LLM can therefore reason about the
environment, decide on a policy in words, and translate it into weights. It can
also read existing weights and edit them, for example, to change the selected
action in a few states. Writing the weights requires neither training data nor
interaction with the environment, but nothing in this process checks the
result: the LLM may misjudge the environment or write wrong weights.
Probabilistic model checking provides this check.

\section{Methodology}\label{sec:method}
Our pipeline, implemented in COOL-MC, takes an MDP $M$, a PCTL property $P_{\sim p}(\rho)$, and a network
architecture, that is, the number of layers and neurons. It returns a policy
network whose weights are all written by an LLM $f$ and verified by Storm. The
pipeline has three steps (Algorithm~\ref{alg:pipeline}): the LLM writes an
initial network (Step~1), optionally repairs it on a dataset of optimal actions
(Step~2), and repairs it with model-checking feedback until the property is
satisfied (Step~3). No step uses gradient descent; we only evaluate the networks
that the LLM writes. All steps query the LLM in the same way, which we describe
first. We write $\pi_N$ for the policy induced by a policy network $N$.

\begin{algorithm}[t]
\caption{Weight-space programming with model-checking feedback.}
\label{alg:pipeline}
\begin{algorithmic}[1]
\Require MDP $M$, property $P_{\sim p}(\rho)$, LLM $f$, network architecture,
  optional description $\delta$ and dataset $\mathcal{D}$, target accuracy
  $\alpha$, round limits $E$ and $R$
\Ensure best verified policy network $N^*$
\State $N \gets \textsc{Request}(\text{creation prompt})$
  \Comment{Step 1}
\For{$e = 1, \ldots, E$} \Comment{Step 2, only if $\mathcal{D}$ is given}
  \State \textbf{if} $\mathit{acc}(N, \mathcal{D}) \geq \alpha$ \textbf{then break}
  \State $N \gets \textsc{Request}(\text{dataset prompt for } N)$
\EndFor
\State $N \gets$ the most accurate network so far
\For{$i = 1, \ldots, R$} \Comment{Step 3}
  \State $q \gets P(\rho)$ in the DTMC induced by $\pi_N$
    \Comment{COOL-MC and Storm}
  \State \textbf{if} $i = 1$ or $q$ is better than $q^*$ \textbf{then}
    $(N^*, q^*) \gets (N, q)$
  \State \textbf{if} $q \sim p$ or $i = R$ \textbf{then break}
  \State $N \gets \textsc{Request}(\text{model-checking prompt for } N)$
\EndFor
\State \Return $N^*$
\Statex
\Function{Request}{$\omega$} \Comment{at most $r$ retries}
  \For{$t = 0, \ldots, r$}
    \State sample $\omega' \sim f(\omega)$ and parse $\omega'$ into weights, notes, and checks
    \If{the answer is invalid}
      \State add the error to $\omega$; \textbf{continue}
    \EndIf
    \If{a check fails and $t < r$}
      \State add the weights and the failed checks to $\omega$; \textbf{continue}
    \EndIf
    \State \Return the network with the parsed weights
  \EndFor
\EndFunction
\end{algorithmic}
\end{algorithm}

\subsection{Querying the LLM}\label{sec:method-query}

\paragraph{Prompts.}
The LLM keeps no memory between calls, so every prompt contains all the
information it needs. Each prompt starts with fixed instructions that explain
how $N$ computes its outputs, recall the ReLU constructions from
Section~\ref{sec:bg-wsp}, and prescribe the answer format. Next comes the
\emph{context} of the task:
\begin{enumerate}
  \item the PRISM code of $M$ and the values of its constants;
  \item the property $P_{\sim p}(\rho)$ and whether a higher or a lower
        probability is better;
  \item the rule that COOL-MC replaces an unavailable action by the first
        available one (Section~\ref{sec:bg-wsp});
  \item the interface of the network: which input is which feature, which
        output is which action, and the shapes of all $W_i$ and $b_i$;
  \item optionally, a natural-language description $\delta$ of the environment.
\end{enumerate}
The prompt ends with the task of the current step. In Steps~2 and~3, it also
contains the current weights, a log of the previous rounds (for example,
accuracies and query results), and the feedback of the step. If the current
network is worse than the best one so far, the prompt also contains the weights
of the best one, so that the LLM can return to it or combine both.

\paragraph{Answers.}
The LLM answers with a JSON object with three fields: the \emph{weights} of all
layers; \emph{notes}, a short explanation of the policy and of what the
neurons compute; and \emph{checks}, 5 to 20 states, each with the action that
the LLM intends its network to select there. Later prompts show the notes
together with the weights, so that the LLM edits its own design instead of
reverse-engineering it from numbers.

\paragraph{Validation and self-check.}
We check every answer in two stages. First, it must be valid JSON with the
required shapes and only finite numbers. Second, the \emph{self-check} evaluates
the written network on the check states and compares its actions with the
intended ones. A mismatch shows that the weights do not encode the LLM's own
policy, for example, because of an arithmetic error or a transposed matrix.
The self-check thus catches encoding errors before Storm runs, whereas Steps~2
and~3 judge the policy itself. If an answer fails either stage, we query the
LLM again with the error added to the prompt. After a failed self-check, we add
the rejected weights and the mismatched states and ask the LLM to correct them
rather than start over. After $r$ retries, we accept the last answer that
passed the first stage. If the creation in Step~1 yields no valid answer at all,
the pipeline stops instead of continuing with random weights. If a request in
Step~2 or~3 yields no valid answer, the network stays unchanged, and Step~2
ends.

\subsection{The Three Steps}\label{sec:method-steps}

\paragraph{Step 1: Creation.}
The prompt contains the environment description $\delta$ as part of the
context, and the task asks the LLM to study the guards, updates, and labels of
the PRISM model, to decide on the best action for each relevant region of the
state space, and to encode this policy in the weights.

\paragraph{Step 2: Dataset repair (optional).}
Storm builds $M$ from $s_0$ and stops exploring at states in which
$\rho = \varphi\ U\ \psi$ is already decided, that is, in which $\psi$ holds or
$\varphi$ is violated. This step uses a dataset
$\mathcal{D} = \{(s, A^*(s)) \mid s \in S_?\}$, where $S_?$ contains the other
states of this model, and $A^*(s) \subseteq \mathit{Act}(s)$ is the set of
\emph{optimal actions} in $s$: an action is optimal in $s$ if taking it and acting
optimally afterwards attains $P_{\max}(\rho)$ (or $P_{\min}(\rho)$) from $s$.
Storm computes $\mathcal{D}$ on $M$. The \emph{accuracy} of $N$ is the fraction of
these states in which the action with the largest output of $N$ is optimal:
\[
  \mathit{acc}(N, \mathcal{D}) =
  \frac{|\{(s, A^*(s)) \in \mathcal{D} \mid
  \operatorname*{arg\,max}_{a \in \mathit{Act}} N(s)_a \in A^*(s)\}|}{|\mathcal{D}|}.
\]
An unavailable action thus counts as wrong.
In each round, the prompt lists the misclassified states, each with its optimal
actions and the action of $N$ (at most 200, randomly sampled if there are more),
and the LLM edits the weights. After at most $E$ rounds, or as soon as the
accuracy reaches $\alpha$, we keep the most accurate network. Accuracy is only a
proxy: it also counts states that $\pi_N$ never reaches, and these do not affect
the property. Step~3 therefore makes the final decision.

\paragraph{Step 3: Model-checking repair.}
In each round, COOL-MC builds the DTMC induced by $\pi_N$, and Storm computes
the \emph{query result} $q = P(\rho)$. For a lower bound ($\geq$, $>$), a higher $q$ is
better; for an upper bound ($\leq$, $<$), a lower one. We keep the best network
$N^*$ and its result $q^*$. If $q$ satisfies the bound ${\sim}\,p$, the pipeline
stops. Otherwise, the prompt reports $q$, $q^*$, and $p$, and shows the
decisions of $N$: the selected action in up to 200 reachable states, starting
with all states in which $N$ selects an unavailable action, followed by a random
sample of the others. It also shows one simulated episode with $\pi_N$ of at
most 60 steps. The feedback does not mark which states cause the violation; the
LLM has to identify them from the PRISM code. The repaired network is verified
in the next round. After at most $R$ rounds, the pipeline returns $N^*$, the
best network that Storm has verified.

\subsection{Limitations}\label{sec:method-limitations}
Our method supports memoryless deterministic policy networks within modeled
PRISM MDP environments, limited by the state space and the number of
transitions~\cite{gross2022coolmc}. The same limits apply to the dataset in
Step~2, because Storm computes the optimal actions on $M$. Since the LLM writes
every weight as a number in its output string, the size of the policy network is
limited by the length of the output string and by how many weights the LLM can
keep track of; we therefore focus on small networks. The output strings vary
between calls, so two runs of the pipeline can return different networks. The
verification result of a returned network, however, is reproducible,
because the network no longer changes once it is written. Finally, the pipeline
guarantees that every returned network is verified, but not that a network
satisfying the property is found within $R$ rounds.

\section{Experiments}\label{sec:experiments}
We evaluate our pipeline with two LLMs on three PRISM environments and answer
four research questions:
\begin{itemize}
  \item[\textbf{RQ1}] How close do the LLM-written policies come to the optimal
        actions and the optimal probability, and what does each step of the
        pipeline contribute?
  \item[\textbf{RQ2}] Do policies programmed for two transporter jobs also
        complete three jobs?
  \item[\textbf{RQ3}] Can the pipeline program a policy for a complex safety
        requirement?
  \item[\textbf{RQ4}] How often does the LLM produce invalid or faulty weights?
\end{itemize}

\subsection{Setup}\label{sec:exp-setup}

\paragraph{Environments.}
Table~\ref{tab:environments} summarizes the environments, their PCTL queries,
and the optimal probabilities, which Storm computes on the MDP.

\emph{Frozen Lake} is a $4 \times 4$ grid in which an agent must reach a goal
tile without falling into one of four holes~\cite{gross2022coolmc}. The only
feature is the position $\mathit{pos} = 4 \cdot \mathit{row} + \mathit{col}$.
The ice is slippery: the agent moves in the intended direction with probability
$0.33$ and otherwise slips to one of the two perpendicular directions. We query
$P(\Diamond\, \mathit{goal})$, the probability of eventually reaching the goal.

\emph{Transporter} is a $5 \times 5$ grid in which a taxi picks up passengers at
one location and drops them off at another~\cite{gross2022coolmc}. Every action
outside the gas station costs one unit of fuel, leaving the gas station refills
the tank to 10 units, and running out of fuel ends the episode. A single job
from the start position already uses the full tank, so completing two jobs
requires refueling at least twice. We query
$P(\Diamond\, \mathit{jobs\_done} = 2)$, the probability of completing two jobs.

\emph{Stock Market} is a simplified market in which a trader starts with a
capital of 5 and no stocks~\cite{gross2022coolmc}. In each step, the trader buys
as many stocks as the capital allows (it holds at most 10; stocks beyond that
are paid for but lost), sells all stocks, or holds.
Afterwards, the buy and sell prices rise by one, fall by one, or stay the same
at random; buying makes a rise more likely than a fall, and selling the
opposite. We query $P(\Diamond\, \mathit{success})$, the probability of
eventually reaching a capital of 50.

\begin{table}[t]
  \centering
  \caption{Environments, PCTL queries, and optimal probabilities computed by
    Storm on the MDP. $d$: number of features; $|\mathit{Act}|$: number of
    actions; $|S|$: number of MDP states that Storm builds for the query, which
    stops at states in which the query is decided; Network: hidden layers
    $\times$ neurons per layer.}
  \label{tab:environments}
  \resizebox{\linewidth}{!}{%
  \begin{tabular}{lrrrcll}
    \toprule
    Environment & $d$ & $|\mathit{Act}|$ & $|S|$ & Network & PCTL query & Optimum \\
    \midrule
    Frozen Lake & 1 & 4 & 17 & $1 \times 16$ & $P(\Diamond\, \mathit{goal})$ & $0.822$ \\
    Transporter (2 jobs) & 10 & 6 & 8{,}560 & $2 \times 32$ & $P(\Diamond\, \mathit{jobs\_done} = 2)$ & $1$ \\
    Transporter (3 jobs) & 10 & 6 & 16{,}496 & $2 \times 32$ & $P(\Diamond\, \mathit{jobs\_done} = 3)$ & $1$ \\
    Stock Market & 5 & 3 & 14{,}551 & $2 \times 32$ & $P(\Diamond\, \mathit{success})$ & $1$ \\
    Transporter (safety) & 10 & 6 & 6{,}880 & $2 \times 32$ & $P(\neg(\mathit{passenger} \wedge \mathit{fuel} < 4)\ U\ \mathit{jobs\_done} = 2)$ & $1$ \\
    \bottomrule
  \end{tabular}}
\end{table}

\paragraph{LLMs.}
We compare DeepSeek-V4-Pro and Kimi-K3.1-Pro, which we access through the cloud
service of Ollama (\texttt{deepseek-v4-pro:cloud} and
\texttt{kimi-k3.1-pro:cloud}).

\paragraph{Pipeline configuration.}
Every run executes all three steps. The dataset of Step~2 contains all optimal
actions for the property and is computed by Storm on the MDP. Each run has at
most $E = 5$ dataset rounds with target accuracy $\alpha = 100\%$, at most
$R = 5$ model-checking rounds, and $r = 2$ retries per request. For each
environment, the LLM receives the
same natural-language description $\delta$, which restates the PRISM model and
contains neither optimal probabilities nor policies. We set the threshold $p$
of each property to its optimal probability (Table~\ref{tab:environments}) minus
$10^{-4}$, which absorbs the numerical error of Storm, so a run succeeds only if
it finds a policy within $10^{-4}$ of the optimum.
We use temperature 0, so the
LLM always selects the most likely next token. Because cloud inference is not
guaranteed to be deterministic, we repeat each experiment three times with the
seeds 1, 2, and 3.

RQ1 needs 18 runs (three environments, two LLMs, three seeds) and RQ3 needs 6
runs (two LLMs, three seeds). RQ2 reuses the networks of RQ1, and RQ4 the logs of
RQ1 and RQ3, so they need no additional LLM calls. Each run makes at most
$(1 + E + R - 1) \cdot (r+1) = 30$ LLM calls: one creation, $E$ dataset
repairs, and $R - 1$ model-checking repairs, each with up to $r$ retries.

\paragraph{Technical setup.}
We ran all experiments in a Docker container with 16~GB of RAM and an Intel
Core i7-12700H processor (20 threads) under Ubuntu 25.10. We use
COOL-MC~\cite{gross2022coolmc} with Storm~\cite{hensel2022probabilistic} 1.12.0
(stormpy 1.12.0) for model checking, Python 3.13, and the \texttt{ollama}
Python package 0.6.3 to query the LLMs, which run on the cloud service of
Ollama. The source code is available at
\url{https://github.com/LAVA-LAB/COOL-MC/tree/wsp_verified}.

\subsection{RQ1: Accuracy and Optimality after Each Step}\label{sec:exp-rq1}
This experiment measures how close the LLM-written policies come to the optimal
policy and what each step contributes. After each step, we measure the accuracy
$\mathit{acc}(N, \mathcal{D})$, that is, the fraction of states in which the
network selects an optimal action (Section~\ref{sec:method-steps}), and the
query result $q$ from the initial state. After Step~1, the LLM has seen no
optimal actions. In Step~2, it sees the states in which its network misses an
optimal action, and in Step~3, it receives model-checking feedback. Accuracy can
stay below 100\% even for an optimal policy, because it also counts states that
the policy never reaches (Table~\ref{tab:rq1}).

\begin{table}[t]
  \centering
  \caption{RQ1: Accuracy and query results after each step. Acc.: mean accuracy
    over the three runs (in \%); $\bar{q}$: mean query result ($\bar{q}^*$:
    best verified result after Step~3); Opt.: runs (of 3) that reach the
    optimal probability up to $10^{-4}$; Calls: mean number of LLM calls per
    run.}
  \label{tab:rq1}
  \resizebox{\linewidth}{!}{%
  \begin{tabular}{llcccccccccc}
    \toprule
    & & \multicolumn{3}{c}{After Step 1} & \multicolumn{3}{c}{After Step 2}
      & \multicolumn{3}{c}{After Step 3} & \\
    \cmidrule(lr){3-5} \cmidrule(lr){6-8} \cmidrule(lr){9-11}
    Environment & LLM & Acc. & $\bar{q}$ & Opt. & Acc. & $\bar{q}$ & Opt.
      & Acc. & $\bar{q}^*$ & Opt. & Calls \\
    \midrule
    Frozen Lake & DeepSeek-V4-Pro & 100.00 & 0.822 & 3/3 & 100.00 & 0.822 & 3/3 & 100.00 & 0.822 & 3/3 & 1.0 \\
     & Kimi-K3.1-Pro & 97.92 & 0.822 & 3/3 & 100.00 & 0.822 & 3/3 & 100.00 & 0.822 & 3/3 & 1.3 \\
    Transporter & DeepSeek-V4-Pro & 59.52 & 1.000 & 3/3 & 79.70 & 1.000 & 3/3 & 79.70 & 1.000 & 3/3 & 8.0 \\
     & Kimi-K3.1-Pro & 98.83 & 1.000 & 3/3 & 100.00 & 1.000 & 3/3 & 100.00 & 1.000 & 3/3 & 4.0 \\
    Stock Market & DeepSeek-V4-Pro & 100.00 & 1.000 & 3/3 & 100.00 & 1.000 & 3/3 & 100.00 & 1.000 & 3/3 & 1.7 \\
     & Kimi-K3.1-Pro & 100.00 & 1.000 & 3/3 & 100.00 & 1.000 & 3/3 & 100.00 & 1.000 & 3/3 & 1.0 \\
    \midrule
    All & DeepSeek-V4-Pro & 86.51 & 0.941 & 9/9 & 93.23 & 0.941 & 9/9 & 93.23 & 0.941 & 9/9 & 3.6 \\
     & Kimi-K3.1-Pro & 98.92 & 0.941 & 9/9 & 100.00 & 0.941 & 9/9 & 100.00 & 0.941 & 9/9 & 2.1 \\
    \bottomrule
  \end{tabular}}
\end{table}

\textbf{Analysis.} All 18 runs reach the optimal probability, all of them
already after Step~1 (Table~\ref{tab:rq1}). Step~3 therefore never had to
repair a network: in every run, the first query result met the threshold. The
created networks of Kimi-K3.1-Pro select an optimal action in 98.9\% of the
states on average; after Step~2, they do so in all states in all nine runs.
DeepSeek-V4-Pro reaches 100\% after Step~1 on Frozen Lake and Stock Market, but
only 59.5\% on the transporter, where two of its three networks select an
optimal action in only 39\% of the states. Step~2 raises one of them to 99.6\%.
For the other, all three answers of the first dataset round have a wrong shape,
so Step~2 ends with the created network (Section~\ref{sec:method-query}). Both
networks show that accuracy is only a proxy: with 39\% accuracy, they still
reach the optimal probability, because the states in which they miss an optimal
action do not affect the property. In these environments, the creation step
alone thus yields optimal policies, Step~2 corrects decisions that do not affect
the property from the initial state, and Step~3 confirms the result; RQ3 shows
a run in which Step~3 is needed. Per run, both LLMs need one or two calls on
Frozen Lake and Stock Market. On the transporter, Kimi-K3.1-Pro needs 4.0 calls
per run on average and DeepSeek-V4-Pro 8.0: no created network reaches 100\%
accuracy, so every run has dataset rounds, and 6 of the 12 calls of
Kimi-K3.1-Pro and 17 of the 24 calls of DeepSeek-V4-Pro return no answer, a
wrong shape, or a network that fails the self-check (RQ4).

\subsection{RQ2: Generalization from Two to Three Jobs}\label{sec:exp-rq2}
A policy that encodes the general rule (refuel when needed, pick up, drop off)
should also complete a third job, whereas a policy tied to the number of jobs
may fail. We take the networks $N^*$ of the transporter runs from RQ1, which
were programmed for two jobs, and model check them without any change on the
transporter with three jobs for $P(\Diamond\, \mathit{jobs\_done} = 3)$. With
three jobs, the policy must also act in states with $\mathit{jobs\_done} = 2$,
in which the two-job episode has already ended. This experiment needs no LLM
calls (Table~\ref{tab:rq2}).

\begin{table}[t]
  \centering
  \caption{RQ2: Transporter networks programmed for two jobs, model checked
    with three jobs.}
  \label{tab:rq2}
  \begin{tabular}{lccc}
    \toprule
    LLM & Optimal for 2 jobs & Optimal for 3 jobs & Mean $P(\Diamond\, \mathit{jobs\_done} = 3)$ \\
    \midrule
    DeepSeek-V4-Pro & 3/3 & 2/3 & 0.667 \\
    Kimi-K3.1-Pro & 3/3 & 2/3 & 0.667 \\
    \bottomrule
  \end{tabular}
\end{table}

\textbf{Analysis.} For each LLM, two of the three networks also reach the optimal
probability with three jobs, although the LLM never saw the three-job model
(Table~\ref{tab:rq2}). The third network of each LLM completes three jobs with
probability 0. Since the number of jobs only determines when an
episode ends, the three-job model behaves like the two-job model until the
second job is done, and all networks complete two jobs with probability 1. The
two failing networks therefore fail in the states with
$\mathit{jobs\_done} = 2$, which the two-job property cannot reveal, because
these states lie after its goal.

\subsection{RQ3: Complex Safety Requirement}\label{sec:exp-rq3}
A taxi that runs out of fuel with a passenger on board strands the passenger. We
therefore require that the taxi completes both jobs and, until then, never
carries a passenger with less than 4 units of fuel:
\[
  P_{\geq 1}\big(\neg(\mathit{passenger} \wedge \mathit{fuel} < 4)\ U\
  \mathit{jobs\_done} = 2\big).
\]

We run the pipeline with this property. In addition, we model check the
transporter networks from RQ1, which were programmed only for
$\Diamond\, \mathit{jobs\_done} = 2$, against the requirement. This shows
whether the requirement changes the programmed behavior (Table~\ref{tab:rq3}).

\begin{table}[t]
  \centering
  \caption{RQ3: Fuel-reserve requirement. RQ1 networks: networks programmed only
    for $\Diamond\, \mathit{jobs\_done} = 2$ that satisfy the requirement; RQ3
    runs: runs (of 3) that satisfy it; $\bar{q}^*$ and Calls as in
    Table~\ref{tab:rq1}; Rounds: mean number of model-checking rounds.}
  \label{tab:rq3}
  \begin{tabular}{lccccc}
    \toprule
    LLM & RQ1 networks & RQ3 runs & $\bar{q}^*$ & Rounds & Calls \\
    \midrule
    DeepSeek-V4-Pro & 2/3 & 3/3 & 1.000 & 1.3 & 11.7 \\
    Kimi-K3.1-Pro & 2/3 & 3/3 & 1.000 & 1.0 & 3.3 \\
    \bottomrule
  \end{tabular}
\end{table}

\textbf{Analysis.} The pipeline programs a policy that satisfies the requirement
in all six runs (Table~\ref{tab:rq3}). Four of the six networks from RQ1,
programmed only for $\Diamond\, \mathit{jobs\_done} = 2$, already satisfy it, and the other
two satisfy it with probability 0.
In five of the six RQ3 runs, the created
network satisfies the requirement. In the remaining run
(DeepSeek-V4-Pro, seed~3), no creation answer passed the self-check. The
accepted network selects the intended action in only 7 of its 15 check states
and satisfies the requirement with probability 0; Step~2 repairs it. The run
with seed~2 is the only one of all 24 runs in which Step~3 repairs a network.
The created network satisfies the requirement with probability 1. Step~2 then
raises the accuracy from 37.9\% to 98.4\%, but the most accurate network
satisfies the requirement with probability 0. A single model-checking repair
restores it: the second model-checking round gives probability 1. This run
shows why Step~3 makes the final decision: a network that selects an optimal
action in 98.4\% of the states can still satisfy the requirement with
probability 0. DeepSeek-V4-Pro needs 11.7 calls per run on average,
Kimi-K3.1-Pro 3.3.

\subsection{RQ4: Generation Errors}\label{sec:exp-rq4}
An answer can fail before Storm sees it (Section~\ref{sec:method-query}): it can
be invalid, or its network can fail the self-check. Over all runs of RQ1 and
RQ3, we count the answers with invalid JSON; the answers with wrong network
shapes or non-finite weights, for instance, a weight matrix with 33 instead of
32 rows; the answers that fail the self-check; the answers accepted despite a
failed self-check after all retries; and the requests that the LLM service did
not answer, for instance, because of a server error (Table~\ref{tab:rq4}).

\begin{table}[t]
  \centering
  \caption{RQ4: Generation errors over all runs of RQ1 and RQ3.}
  \label{tab:rq4}
  \resizebox{\linewidth}{!}{%
  \begin{tabular}{lcccccc}
    \toprule
    LLM & Answers & Invalid JSON & Wrong shape & Failed self-check & Accepted despite failure & Failed requests \\
    \midrule
    DeepSeek-V4-Pro & 67 & 0 & 32 & 14 & 5 & 0 \\
    Kimi-K3.1-Pro & 28 & 0 & 8 & 1 & 1 & 1 \\
    \bottomrule
  \end{tabular}}
\end{table}

\textbf{Analysis.} No answer contains invalid JSON, so both LLMs follow the
answer format (Table~\ref{tab:rq4}). The most common error is a wrong network
shape, in 32 of the 67 answers of DeepSeek-V4-Pro (48\%) and in 8 of the 28
answers of Kimi-K3.1-Pro (29\%). In 38 of these 40 answers, a hidden-layer
weight matrix has 30, 31, 33, or 34 rows instead of 32, which suggests that the
LLM loses count while writing a long matrix. The other two answers have a wrong
number of layers.
DeepSeek-V4-Pro also fails the self-check far more often,
and five of its networks were accepted despite a failed self-check after all
retries, compared to one of Kimi-K3.1-Pro. One request to Kimi-K3.1-Pro timed
out. No run failed because of these errors, but they cost calls:
DeepSeek-V4-Pro sent 67 requests for its 12 runs, Kimi-K3.1-Pro 29 (28 answers
and the timed-out request).

\section{Discussion}\label{sec:discussion}

\paragraph{Benefits.}
The creation step produces a neural network policy without gradient decent,
without interaction with the environment, and without a reward function. The LLM
works from the PRISM model and the PCTL property, so it can target requirements
that rewards cannot express, such as the until requirement of RQ3. In all 18
runs of RQ1, the creation step yields an optimal policy. The key
benefit, however, is that every returned network is verified. Neither the LLM
nor the dataset reveals all failures: in RQ3, dataset repair broke a policy that
had satisfied the requirement, and in RQ2, two policies that are optimal for two
jobs fail at the third. Model checking exposes such failures, quantifies them
by a probability, and, in Step~3, turns them into feedback. Because the LLM writes
ordinary weights, the result remains a standard neural network policy, to which
existing tools for neural policies apply, for instance, verification with
COOL-MC~\cite{gross2022coolmc}.

\paragraph{Future applications.}
Step~3 only needs an MDP, a network, and a property, so it could also repair
policies trained with RL: when a trained policy violates a property or a
requirement changes, the LLM edits a few weights instead of retraining the
network, and Storm confirms the repair. A verified LLM-written policy could also
serve as a starting point for RL training or as a safe fallback policy.
Finally, RQ2 suggests verifying a written policy on a family of models, for
example, with different numbers of jobs, and feeding the failures back, so that
the LLM encodes general rules instead of solving a single instance.

Our setting can also serve as a benchmark for the weight-space programming
skills of LLMs. Storm provides the ground truth: the optimal probability and
the optimal actions. The difficulty can be scaled through the
size of the environment, the size of the network, and the complexity of the
property, and changing the constants of a PRISM model yields new variants. Our results already separate the two
LLMs: both reach the optimal probabilities, but they differ in accuracy, in
generation errors, and in the number of calls they need (RQ1, RQ4).

Weight-space programming could also support safety explanations, that is,
explanations of which parts of a policy are responsible for which safety
property. Existing methods analyze trained networks after the fact, for
instance, by pruning connections and measuring how the query results
change~\cite{gross2024safety} or by analyzing which neurons are activated
together~\cite{gross2025co}. In a written network, the LLM's notes
already state which neurons are meant to compute which rule, for example, a neuron that
sends the taxi to the gas station while it carries a passenger. Disabling such
a neuron and model checking the network again shows which properties depend on
it: if the fuel-reserve requirement of RQ3 is then violated, the neuron is
responsible for it.
An LLM could read the
weights of a trained network and point to the neurons responsible for a safety
property, and model checking would confirm or refute its claim.

\paragraph{Threats to validity.}
Each configuration runs three times, and cloud inference is not deterministic
even at temperature 0, so our numbers are indicative rather than precise
estimates; the cloud models may also change over time. The environments and
networks are small (at most 16{,}496 states and two hidden layers with 32
neurons), and we test two LLMs, one generalization task, and one complex safety
requirement, so the results may not transfer to larger environments, larger
networks, or other LLMs. The PRISM models are publicly available with COOL-MC,
and Frozen Lake is a well-known benchmark, so we cannot rule out that the LLMs
saw these models or related policies during training. We wrote the
environment descriptions ourselves; they contain no policies, but they restate
the model, and other
descriptions could change the results.

\section{Conclusion and Future Work}\label{sec:conclusion}
We presented a pipeline in which an LLM writes the weights of a neural network
policy for a PRISM environment and repairs them, guided by a self-check, a
dataset of optimal actions, and probabilistic model checking with COOL-MC and
Storm. No step uses gradient descent, and Storm verifies every returned
network. With two LLMs on three environments, the created policy of each of the
18 runs reaches the optimal probability. Four of the six policies programmed for
two transporter jobs also complete three jobs with probability 1, a task that the
LLM never saw, and the pipeline programs a fuel-reserve requirement, expressed with the until
operator, in all six runs. In one of these runs, dataset repair raised the
accuracy to 98.4\% but broke the requirement, which model checking revealed,
and a single model-checking repair restored it. The most common generation error, a weight
matrix with a wrong number of rows, is caught before model checking. Weight-space
programming and probabilistic model checking thus complement each other: the
LLM writes policies without training, and model checking computes how well each
of them satisfies the property.

In future work, we will test the repair more thoroughly: 23 of the 24 created
policies already satisfied their property, so model-checking repair was rarely
needed. Harder environments and requirements,
in which the creation step fails, would show how well the repair works.
Richer feedback, for example, the states that cause the violation, could make
the repairs more targeted. The wrong shapes in RQ4 suggest that LLMs lose count
when writing long matrices; writing the weights neuron by neuron, in sparse
form, or through generated code could scale the pipeline to larger networks.
Finally, we plan to evaluate the applications from
Section~\ref{sec:discussion}: repairing policies trained with RL, safety
explanations, and benchmarking further LLMs.

\bibliographystyle{unsrt}  
\bibliography{references}

@article{mnih2015human,
  author    = {Volodymyr Mnih and
               Koray Kavukcuoglu and
               David Silver and
               Andrei A. Rusu and
               Joel Veness and
               Marc G. Bellemare and
               Alex Graves and
               Martin A. Riedmiller and
               Andreas Fidjeland and
               Georg Ostrovski and
               Stig Petersen and
               Charles Beattie and
               Amir Sadik and
               Ioannis Antonoglou and
               Helen King and
               Dharshan Kumaran and
               Daan Wierstra and
               Shane Legg and
               Demis Hassabis},
  title     = {Human-level control through deep reinforcement learning},
  journal   = {Nat.},
  volume    = {518},
  number    = {7540},
  pages     = {529--533},
  year      = {2015}
}

@article{hansson1994logic,
  title={A logic for reasoning about time and reliability},
  author={Hansson, Hans and Jonsson, Bengt},
  journal={Formal aspects of computing},
  volume={6},
  number={5},
  pages={512--535},
  year={1994},
  publisher={Springer}
}

@inproceedings{gross2022coolmc,
  title={COOL-MC: a comprehensive tool for reinforcement learning and model checking},
  author={Gross, Dennis and Jansen, Nils and Junges, Sebastian and P{\'e}rez, Guillermo A},
  booktitle={International Symposium on Dependable Software Engineering: Theories, Tools, and Applications},
  pages={41--49},
  year={2022},
  organization={Springer}
}

@article{huang2024survey,
  title={A survey of safety and trustworthiness of large language models through the lens of verification and validation},
  author={Huang, Xiaowei and Ruan, Wenjie and Huang, Wei and Jin, Gaojie and Dong, Yi and Wu, Changshun and Bensalem, Saddek and Mu, Ronghui and Qi, Yi and Zhao, Xingyu and others},
  journal={Artificial Intelligence Review},
  volume={57},
  number={7},
  pages={175},
  year={2024},
  publisher={Springer}
}

@inproceedings{kwiatkowska2011prism,
  title={PRISM 4.0: Verification of probabilistic real-time systems},
  author={Kwiatkowska, Marta and Norman, Gethin and Parker, David},
  booktitle={International conference on computer aided verification},
  pages={585--591},
  year={2011},
  organization={Springer}
}

@book{baier2008principles,
  title     = {Principles of model checking},
  author    = {Baier, Christel and Katoen, Joost-Pieter},
  year      = {2008},
  publisher = {MIT press}
}

@article{hensel2022probabilistic,
  title={The probabilistic model checker Storm},
  author={Hensel, Christian and Junges, Sebastian and Katoen, Joost-Pieter and Quatmann, Tim and Volk, Matthias},
  journal={International Journal on Software Tools for Technology Transfer},
  volume={24},
  number={4},
  pages={589--610},
  year={2022},
  publisher={Springer}
}

@article{han2026survey,
  title={A survey of weight space learning: Understanding, representation, and generation},
  author={Han, Xiaolong and Wang, Zehong and Zhao, Bo and Zhang, Binchi and Li, Jundong and Borth, Damian and Yu, Rose and Maron, Haggai and Ye, Yanfang and Yin, Lu and others},
  journal={arXiv preprint arXiv:2603.10090},
  year={2026}
}

@article{DBLP:journals/aamas/VamplewSKRRRHHM22,
  author    = {Peter Vamplew and
               Benjamin J. Smith and
               Johan K{\"{a}}llstr{\"{o}}m and
               Gabriel de Oliveira Ramos and
               Roxana Radulescu and
               Diederik M. Roijers and
               Conor F. Hayes and
               Fredrik Heintz and
               Patrick Mannion and
               Pieter J. K. Libin and
               Richard Dazeley and
               Cameron Foale},
  title     = {Scalar reward is not enough: a response to Silver, Singh, Precup and
               Sutton {(2021)}},
  journal   = {Auton. Agents Multi Agent Syst.},
  volume    = {36},
  number    = {2},
  pages     = {41},
  year      = {2022}
}

@article{huynh2025large,
  title={Large language models for code generation: A comprehensive survey of challenges, techniques, evaluation, and applications},
  author={Huynh, Nam and Lin, Beiyu},
  journal={arXiv preprint arXiv:2503.01245},
  year={2025}
}

@article{vaswani2017attention,
  title={Attention is all you need},
  author={Vaswani, Ashish and Shazeer, Noam and Parmar, Niki and Uszkoreit, Jakob and Jones, Llion and Gomez, Aidan N and Kaiser, {\L}ukasz and Polosukhin, Illia},
  journal={Advances in neural information processing systems},
  volume={30},
  year={2017}
}

@article{gross2025verifying,
  title={Verifying Memoryless Sequential Decision-making of Large Language Models},
  author={Gross, Dennis and Spieker, Helge and Gotlieb, Arnaud},
  journal={arXiv preprint arXiv:2510.06756},
  year={2025}
}

@article{Yang2022safety-constrained,
  title = {Safety-constrained reinforcement learning with a distributional safety critic},
  author = {Yang, Qisong and Sim{\~a}o, Thiago D. and Tindemans, Simon H. and Spaan, Matthijs T. J.},
  journal = {Machine Learning},
  volume = {112},
  number = {3},
  pages = {859--887},
  year = {2022},
  publisher = {Springer}
}

@inproceedings{azeem20251,
  title={1--2--3--go! policy synthesis for parameterized markov decision processes via decision-tree learning and generalization},
  author={Azeem, Muqsit and Chakraborty, Debraj and Kanav, Sudeep and K{\v{r}}et{\'\i}nsk{\`y}, Jan and Mohagheghi, Mohammadsadegh and Mohr, Stefanie and Weininger, Maximilian},
  booktitle={International Conference on Verification, Model Checking, and Abstract Interpretation},
  pages={97--120},
  year={2025},
  organization={Springer}
}

@article{gross2026formally,
  title={Formally Verifying and Explaining Sepsis Treatment Policies with COOL-MC},
  author={Gross, Dennis},
  journal={arXiv preprint arXiv:2602.14505},
  year={2026}
}

@inproceedings{ashok2021dtcontrol,
  title={dtControl 2.0: explainable strategy representation via decision tree learning steered by experts},
  author={Ashok, Pranav and Jackermeier, Mathias and K{\v{r}}et{\'\i}nsk{\`y}, Jan and Weinhuber, Christoph and Weininger, Maximilian and Yadav, Mayank},
  booktitle={International Conference on Tools and Algorithms for the Construction and Analysis of Systems},
  pages={326--345},
  year={2021},
  organization={Springer}
}

@inproceedings{DBLP:conf/tacas/HahnPSSTW19,
  author    = {Ernst Moritz Hahn and
               Mateo Perez and
               Sven Schewe and
               Fabio Somenzi and
               Ashutosh Trivedi and
               Dominik Wojtczak},
  title     = {Omega-Regular Objectives in Model-Free Reinforcement Learning},
  booktitle = {{TACAS} {(1)}},
  series    = {{LNCS}},
  volume    = {11427},
  pages     = {395--412},
  publisher = {Springer},
  year      = {2019}
}

@article{Konighofer2025Shields,
  author = {K{\"o}nighofer, Bettina and Bloem, Roderick and Jansen, Nils and Junges, Sebastian and Pranger, Stefan},
  title = {Shields for Safe Reinforcement Learning},
  journal = {Communications of the ACM},
  volume = {68},
  number = {11},
  pages = {80--90},
  year = {2025}
}

@inproceedings{brorholt2026adaptive,
  title={Adaptive Probabilistic Shielding by Learning MDPs for Safe Reinforcement Learning},
  author={Brorholt, Astrid Horn and Galesloot, Maris FL and Jansen, Nils and Larsen, Kim Guldstrand and Schilling, Christian},
  booktitle={International Conference on Runtime Verification},
  pages={41--60},
  year={2026},
  organization={Springer}
}

@inproceedings{verma2018programmatically,
  title={Programmatically interpretable reinforcement learning},
  author={Verma, Abhinav and Murali, Vijayaraghavan and Singh, Rishabh and Kohli, Pushmeet and Chaudhuri, Swarat},
  booktitle={International conference on machine learning},
  pages={5045--5054},
  year={2018},
  organization={PMLR}
}

@inproceedings{ma2024eureka,
  title={Eureka: Human-level reward design via coding large language models},
  author={Ma, Yecheng Jason and Liang, William and Wang, Guanzhi and Huang, De-An and Bastani, Osbert and Jayaraman, Dinesh and Zhu, Yuke and Fan, Jim and others},
  booktitle={International conference on learning Representations},
  volume={2024},
  pages={26516--26560},
  year={2024}
}

@inproceedings{liang2023code,
  title={Code as policies: Language model programs for embodied control},
  author={Liang, Jacky and Huang, Wenlong and Xia, Fei and Xu, Peng and Hausman, Karol and Ichter, Brian and Florence, Pete and Zeng, Andy},
  booktitle={2023 IEEE International conference on robotics and automation (ICRA)},
  pages={9493--9500},
  year={2023},
  organization={IEEE}
}

@inproceedings{weiss2021thinking,
  title={Thinking like transformers},
  author={Weiss, Gail and Goldberg, Yoav and Yahav, Eran},
  booktitle={International Conference on Machine Learning},
  pages={11080--11090},
  year={2021},
  organization={PMLR}
}

@inproceedings{NEURIPS2023_771155ab,
 author = {Lindner, David and Kramar, Janos and Farquhar, Sebastian and Rahtz, Matthew and McGrath, Tom and Mikulik, Vladimir},
 booktitle = {Advances in Neural Information Processing Systems},
 doi = {10.52202/075280-1649},
 editor = {A. Oh and T. Naumann and A. Globerson and K. Saenko and M. Hardt and S. Levine},
 pages = {37876--37899},
 publisher = {Curran Associates, Inc.},
 title = {Tracr: Compiled Transformers as a Laboratory for Interpretability},
 url = {https://proceedings.neurips.cc/paper_files/paper/2023/file/771155abaae744e08576f1f3b4b7ac0d-Paper-Conference.pdf},
 volume = {36},
 year = {2023}
}

@article{meng2022locating,
  title={Locating and editing factual associations in gpt},
  author={Meng, Kevin and Bau, David and Andonian, Alex and Belinkov, Yonatan},
  journal={Advances in neural information processing systems},
  volume={35},
  pages={17359--17372},
  year={2022}
}

@article{ilharco2022editing,
  title={Editing models with task arithmetic},
  author={Ilharco, Gabriel and Ribeiro, Marco Tulio and Wortsman, Mitchell and Gururangan, Suchin and Schmidt, Ludwig and Hajishirzi, Hannaneh and Farhadi, Ali},
  journal={arXiv preprint arXiv:2212.04089},
  year={2022}
}

@inproceedings{zeng2026generative,
  title={Generative modeling of weights: Generalization or memorization?},
  author={Zeng, Boya and Yin, Yida and Xu, Zhiqiu and Liu, Zhuang},
  booktitle={Proceedings of the IEEE/CVF Conference on Computer Vision and Pattern Recognition},
  pages={41974--41984},
  year={2026}
}

@article{charakorn2025text,
  title={Text-to-lora: Instant transformer adaption},
  author={Charakorn, Rujikorn and Cetin, Edoardo and Tang, Yujin and Lange, Robert Tjarko},
  journal={arXiv preprint arXiv:2506.06105},
  year={2025}
}

@article{peebles2022learning,
  title={Learning to learn with generative models of neural network checkpoints},
  author={Peebles, William and Radosavovic, Ilija and Brooks, Tim and Efros, Alexei A and Malik, Jitendra},
  journal={arXiv preprint arXiv:2209.12892},
  year={2022}
}

@article{zhou2023permutation,
  title={Permutation equivariant neural functionals},
  author={Zhou, Allan and Yang, Kaien and Burns, Kaylee and Cardace, Adriano and Jiang, Yiding and Sokota, Samuel and Kolter, J Zico and Finn, Chelsea},
  journal={Advances in neural information processing systems},
  volume={36},
  pages={24966--24992},
  year={2023}
}

@inproceedings{navon2023equivariant,
  title={Equivariant architectures for learning in deep weight spaces},
  author={Navon, Aviv and Shamsian, Aviv and Achituve, Idan and Fetaya, Ethan and Chechik, Gal and Maron, Haggai},
  booktitle={International Conference on Machine Learning},
  pages={25790--25816},
  year={2023},
  organization={PMLR}
}

@article{DBLP:journals/corr/HaDL16,
  author       = {David Ha and
                  Andrew M. Dai and
                  Quoc V. Le},
  title        = {HyperNetworks},
  journal      = {CoRR},
  volume       = {abs/1609.09106},
  year         = {2016},
  url          = {https://doi.org/10.48550/arXiv.1609.09106},
  doi          = {10.48550/ARXIV.1609.09106},
  eprinttype   = {arXiv},
  eprint       = {1609.09106},
  bibsource    = {dblp computer science bibliography, https://dblp.org}
}

@article{xu2026deepseek,
  title={Deepseek-v4: Towards highly efficient million-token context intelligence},
  author={Xu, Anyi and Lin, Bangcai and Xue, Bing and Wang, Bingxuan and Xu, Bingzheng and Wu, Bochao and Zhang, Bowei and Lin, Chaofan and Dong, Chen and Ling, Chenchen and others},
  journal={arXiv preprint arXiv:2606.19348},
  year={2026}
}

@article{gross2025co,
  title={Co-activation graph analysis of safety-verified and explainable deep reinforcement learning policies},
  author={Gross, Dennis and Spieker, Helge},
  journal={arXiv preprint arXiv:2501.03142},
  year={2025}
}

@article{takehi2026sparks,
  title={Sparks of Weight-Space Programming: LLMs can natively read \& write neural network weights},
  author={Takehi, Rikiya and Teoh, Jayden and Zhang, Alex and Khattab, Omar},
  journal={Research blog, November},
  volume={2},
  year={2026}
}

@article{team2026kimi,
  title={Kimi k3: Open frontier intelligence},
  author={Team, Kimi and Bai, Tongtong and Bai, Yifan and Bao, Yiping and Cai, Jianfeng and Cai, Xinyuan and Cao, Peizhou and Cao, Yuxuan and Chai, Ziwei and Charles, Y and others},
  journal={arXiv preprint arXiv:2607.24653},
  year={2026}
}

@article{gross2024safety,
  title={Safety-oriented pruning and interpretation of reinforcement learning policies},
  author={Gross, Dennis and Spieker, Helge},
  journal={arXiv preprint arXiv:2409.10218},
  year={2024}
}

@article{cha2026consistency,
  title={Consistency Is Not Understanding: Orbit Audits for Neural Field Weight Readers},
  author={Cha, Sungguk and Kim, Mintae and Han, Youngsub and Jeon, Byoung-Ki and Lee, Sangyeob},
  journal={arXiv preprint arXiv:2611.05798},
  year={2026}
}

@article{zheng2026weight,
  title={Weight-Space Programming for Improving Image-Restoration Diffusion Models: Technical Conjectures and a Recursive Improvement Loop},
  author={Zheng, Zhuoran and others},
  journal={arXiv preprint arXiv:2611.04268},
  year={2026}
}

\end{document}